# YOLOSCM:An improved YOLO algorithm for cars detection

**Changhui Deng[1,4], Lieyang Chen[2,5], Shinan Liu[3,6]**

[1] School of Computer Science, Hubei University, Hubei, China
[2] Columbia University in the City of New York, New York, USA
[3] College of engineering, Northeastern University, Boston, USA

[4] 1335372109@qq.com
[5] lc3548@columbia.edu
[6] Liu.shin@northeastern.edu

**Abstract.** Detecting objects in urban traffic images presents considerable difficulties because of the following reasons: 1) These images are typically immense in size, encompassing millions or even hundreds of millions of pixels, yet computational resources are constrained. 2) The small size of vehicles in certain scenarios leads to insufficient information for accurate detection. 3) The uneven distribution of vehicles causes inefficient use of computational resources. To address these issues, we propose YOLOSCM (You Only Look Once with Segmentation Clustering Module), an efficient and effective framework. To address the challenges of large-scale images and the non-uniform distribution of vehicles, we propose a Segmentation Clustering Module (SCM). This module adaptively identifies clustered regions, enabling the model to focus on these areas for more precise detection. Additionally, we propose a new training strategy to optimize the detection of small vehicles and densely packed targets in complex urban traffic scenes. We perform extensive experiments on urban traffic datasets to demonstrate the effectiveness and superiority of our proposed approach.

## 1. Introduction

Object detection has advanced significantly in recent years, driven by the rapid evolution of deep learning techniques. Detectors like Faster R-CNN, YOLO, and SSD have demonstrated remarkable performance on natural image datasets such as MS COCO and Pascal VOC. Despite these achievements, their accuracy and efficiency remain inadequate for vehicle detection in urban traffic images.

Urban traffic images, typically captured by cameras on vehicles or surveillance systems, provide a rich visual field but also present unique challenges for object detection. Vehicle detection in these images is more challenging for three main reasons: 1) Urban traffic images are often very large, containing a large number of vehicles, and require efficient processing strategies to handle the scale of the image data. 2) Urban traffic scenes contain many small and distant vehicles, making it difficult for detectors to recognize them due to limited resolution and visual features. Small vehicles are generally defined as those occupying fewer than $32 \times 32$ pixels. 3) The distribution of vehicles in urban traffic images is highly non-uniform. For example, as shown in Fig. 1, vehicles may cluster at traffic lights or intersections, while others may be sparse. Vehicles in congested regions require more focus, while areas with no vehicles should be disregarded.

Early studies [1] introduced innovative multi-scale training strategies, such as scale normalization for image pyramids (SNIP) and its improved versions. While these methods effectively enhance the detection of small vehicles, they demand substantial computational resources and memory. Another line of research focuses on increasing image or feature resolution. For example, generative adversarial networks (GANs) can recover details lost in small vehicle features, reducing the disparity between small and large vehicle representations. However, these approaches are computationally expensive.

Recently, label-assignment-based methods [2] have been proposed to optimize the assignment of rare small vehicle samples, improving detection performance. Despite these advancements, there is still considerable scope for enhancing both accuracy and efficiency.

The uneven distribution of vehicles in urban traffic images poses additional challenges, hindering the efficiency and accuracy of detectors on large-scale traffic datasets. A straightforward solution is to crop images into smaller sections and scale them up, as done in uniform cropping. However, this method overlooks the non-uniform distribution of vehicles and requires significant computational effort to process all cropped regions.

To address the above technical challenges, recent solutions have focused on designing specific schemes to identify vehicle cluster regions [3] that can then be used for detection. For example, ClusDet [4] uses a clustered detection network to detect vehicle clusters, while DMNet [5] generates cluster regions through density maps by modeling vehicle distribution. These methods have shown promising results by preserving clustered regions and suppressing background as much as possible. However, independently detecting each crop can slow down inference speed. Furthermore, although these methods generate cluster regions, some clusters may have sparse vehicle distributions, resulting in little improvement in final detection performance. Therefore, finding the optimal balance between accuracy and efficiency remains a critical challenge for vehicle detection in urban traffic images.

In this paper, aiming at the challenge of small vehicle detection in urban traffic scenes, we enhance the recognition ability of the model for low-resolution and small-size vehicles by optimizing the structure of the detection head. At the same time, we also improve the training strategy, adopting multi-stage learning and data enhancement techniques to further improve the generalization ability of the model. To further improve detection accuracy, we combine an unsupervised clustering method and use the clustering results as prior knowledge in the training process, thus effectively improving the model's performance in detecting small vehicles. The experimental results show that the proposed method can significantly improve the precision of small vehicle detection tasks and better adapt to different traffic scenarios and target size variations.

## 2. METHODOLOGY

### 2.1. Small Target Feature Map

One of the reasons for the poor performance of YOLOv8 in vehicle detection, particularly for small vehicles, is that the size of small vehicle samples is small, and the downsampling factor in YOLOv8 is relatively large. This makes it difficult for deep feature maps to learn the feature information of small vehicles. To address this, we propose adding a small vehicle detection layer to detect shallow feature maps after they are fused with deep feature maps. The addition of the small vehicle detection layer allows the network to focus more on detecting small vehicles and improves the overall detection performance.

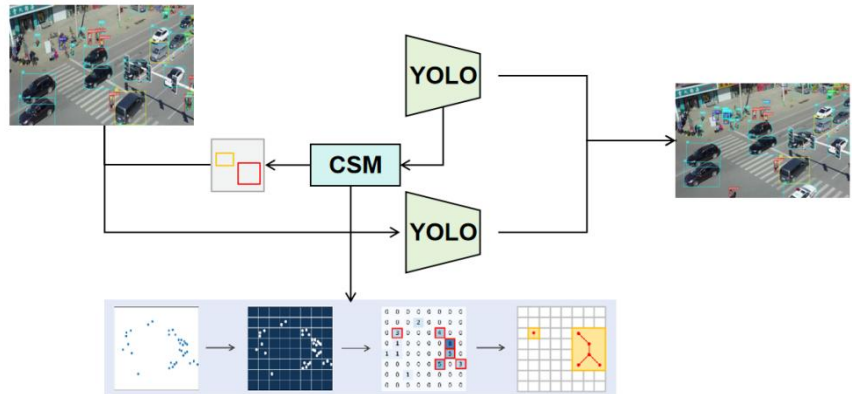

**Figure 1. Architecture of purposed method**

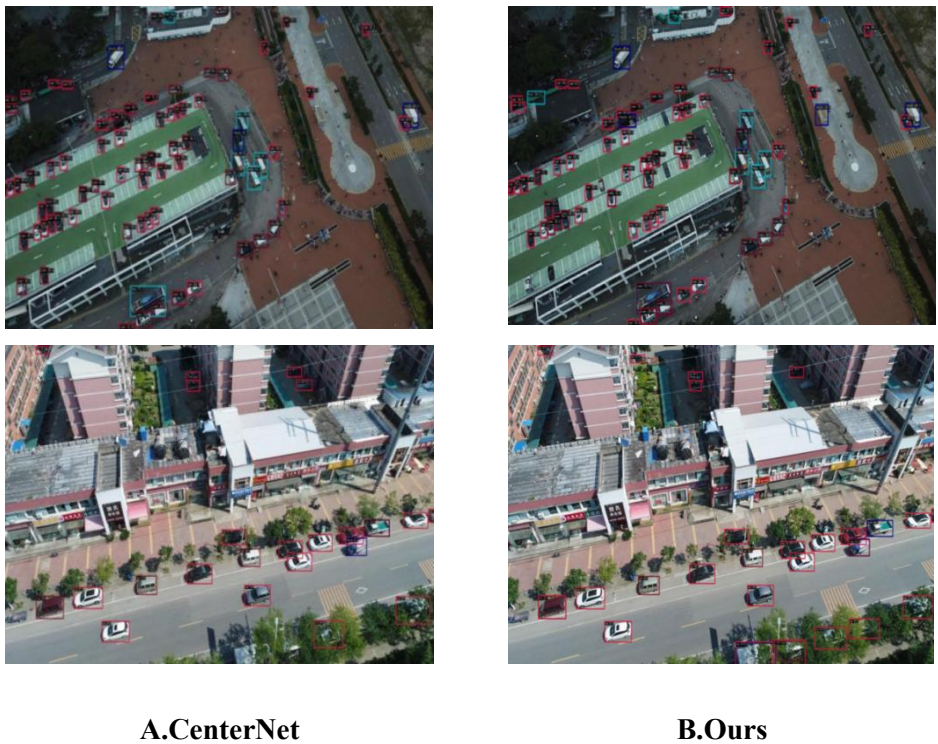

        **A.CenterNet**                            **B.Ours**

**Figure2.** Results between CenterNet and our method

*2.2. Improved joint learning strategies*

    In our training method, a phased training strategy is adopted to optimize the detection of small vehicles and densely distributed vehicles in complex urban traffic scenarios.

    Rough training phase (target aggregation on large-scale images): In the first stage of training, we use large-scale input images for rough training. During this phase, the model focuses on identifying the general areas where vehicles are aggregated across a large portion of the image, rather than focusing on individual small vehicles. By using a lower resolution or scaled image, the model can

quickly learn the general position and shape of vehicles and produce a preliminary feature map. In this stage, the model detects a series of potential vehicle regions, but these regions are often rough and may contain overlapping or closely packed vehicles.

Clustering segmentation module: To further improve detection accuracy, we introduce a clustering segmentation module to finely segment the feature maps obtained from the rough training stage. This module can effectively extract the boundaries between different vehicles and eliminate redundant regions. After clustering, each vehicle region is clearly separated, laying the foundation for subsequent fine training. The details of the clustering module are covered in the next section.

Fine training stage (segmented images for further training): In the second stage, the model undergoes fine training using images processed by the clustering segmentation module. At this point, the image has been subdivided into multiple independent vehicle regions, allowing the model to focus on learning the detailed features of each vehicle, such as edges, subtle variations, and feature points. Compared to the rough training stage, the goal of the fine training stage is to improve detection accuracy, especially in the positioning and classification of small vehicles. At this stage, the model makes more accurate boundary predictions and classifications for each vehicle, resulting in higher precision detection [6].

Through this phased training strategy, the vehicle is first roughly identified and positioned, then refined through clustering segmentation, and finally, small and densely packed vehicles are accurately detected. This method ensures efficient training while significantly improving the detection accuracy of the model, especially for vehicle detection tasks in complex traffic scenarios.

*2.3. Segmentation Clustering module*

Region of Interest (ROI) proposal methods play a critical role in cropping-based object detection models. In urban traffic images, vehicles often cluster in specific areas, leaving much of the image as irrelevant background. Additionally, the limited resolution in dense vehicle regions leads to a significant decline in detection performance. Existing cropping-based methods, such as DMNet, generate excessive crops or rely on additional networks like ClusDet, resulting in slower detection speeds and increased model complexity.

To address these challenges, we propose a Segmentation Clustering Module (SCM) that adaptively identifies dense vehicle regions. Drawing inspiration from YOLC [7], we introduce key modifications tailored for urban traffic images. First, rather than detecting a single maximum clustered region, SCM identifies the top-K densest regions by ranking grid densities, accommodating multiple vehicle clusters commonly found in such images. Second, while AutoScale performs well in single-class crowd localization [8], it struggles with the multi-category nature of urban traffic. Similarly, clustering-based methods like DBSCAN and K-Means, used in UCGNet [9], often generate oversized crops that ignore density variations.

The SCM operates as follows:

1) A binary vehicle presence map is created by generating a heatmap and binarizing it with a threshold to produce a location mask.

2) The heatmap is divided into grids, and the top-K densest grids are selected as candidates. Optimal parameters for urban traffic datasets are Grid = (16, 10) and top-K = 30.

3) To ensure complete vehicle coverage, grids connected by eight neighbors are merged into larger candidate regions.

4) Dense regions are extracted and resized to fit the detection model.

To enhance efficiency and performance, SCM limits the number of crops to k=2 per image, focusing on larger dense regions, as fine detection on these regions yields better results. Experiments show performance stabilizes when k exceeds 3, making this approach both effective and efficient. SCM, as an unsupervised module, integrates seamlessly into any keypoint-based vehicle detection framework, providing a balance between detection speed and accuracy.

## 3. Experiments and Analysis

*3.1. Dataset*

The VisDrone 2019 dataset is a large-scale, high-quality resource designed for drone-based object detection, tracking, and recognition tasks. It was created for the VisDrone 2019 Challenge, part of the International Conference on Computer Vision (ICCV) 2019.

The dataset consists of images captured from a variety of real-world drone flight scenarios, including urban streets, rural areas, and highways. This diversity makes it ideal for developing and evaluating robust vision systems that can operate in different environments. Annotated with a wide range of object categories commonly encountered in drone surveillance—such as pedestrians, cars, bicycles, buses, trucks, and other objects—it is well-suited for both detection and tracking tasks. The dataset contains over 10,000 images and more than 40,000 annotated object instances. It is divided into training, validation, and test sets, with over 10,000 bounding boxes in the training set.

Each object in the dataset is labeled with a bounding box and a category. In addition, the tracking dataset includes object ID annotations, enabling multi-object tracking tasks.

The images in the dataset are high-resolution, captured from various altitudes, providing detailed views of the objects within the scenes. This high level of detail is especially beneficial for developing algorithms capable of handling small objects at greater altitudes. The dataset poses several challenges, including varying object scales, occlusions, complex backgrounds, and changes in lighting conditions. It thus serves as an excellent real-world testbed for evaluating object detection, tracking, and multi-task learning models.The VisDrone dataset is widely used in research and competitions, providing a benchmark for evaluating the performance of computer vision algorithms, particularly in the context of autonomous drone systems and surveillance applications.

*3.2. Evaluation Metrics*

Following the evaluation protocol of the MS COCO dataset, we use AP, AP50, and AP75 as our primary metrics. AP (Average Precision) measures the mean precision across all categories, while AP50 and AP75 represent the average precision at Intersection over Union (IoU) thresholds of 0.5 and 0.75, respectively. Additionally, we report the average precision for each object category individually to assess the model's per-class performance. To evaluate the efficiency of our approach, we record the processing time per GPU for a single original image.

*3.3. Implementation Details*

The model undergoes training for 200 epochs utilizing the SGD optimizer, configured with a momentum of 0.9 and a weight decay of 0.0001. The training process is performed on a GeForce RTX 4080 GPU with a batch size of 4. An initial learning rate of 0.01 is adopted, incorporating a linear warm-up strategy. The input resolution is consistently set at 1024 × 640 across both datasets.Results Analysis

Table I presents a quantitative comparison with state-of-the-art methods on the VisDrone dataset. The results show that our proposed model consistently outperforms other methods on this dataset. Notably, general object detectors like CenterNet [10] perform poorly due to the large number of small vehicle instances and the non-uniform data distribution typical of aerial images.

Our method excels in detecting vehicles with non-uniform distribution, as demonstrated in the third image of the first row. Furthermore, in Fig. 2, we compare the performance of YOLOSCM and CenterNet, showing that YOLOSCM achieves more accurate detections, particularly for small vehicles in dense regions, when compared to CenterNet.

**Table I.** Comparison different algorithms on the VisDrone2019(cars).

| Method | AP | AP50 | AP75 |
|---|---|---|---|
| ClusDet | 23.6 | 31.5 | 22.5 |
| CDMNet | 26.7 | 34.2 | 28.6 |
| GLSAN | 27.0 | 33.1 | 28.7 |
| CenterNet | 23.1 | 31.8 | 21.8 |
| Ours | 28.2 | 35.5 | 29.7 |

## 4. Conclusion

This paper introduces an anchor-free YOLOSCM framework, designed to address two major challenges in vehicle detection from aerial images: detecting small vehicles and handling uneven data distribution. To address the first challenge, a high-resolution feature branch is added to the network, enabling better detection of small vehicles. To solve the second issue, a novel training strategy is proposed. During rough training, the locations of target aggregation are identified, and the Segmentation Clustering Module (SCM) is then used to retain only those regions containing vehicles for fine detection. This approach significantly improves detection performance, especially for small vehicles.Our framework strikes a balance between accuracy and speed through the integration of SCM. The effectiveness and superiority of this method are demonstrated through experiments on the VisDrone2019 dataset, where our approach is compared with state-of-the-art methods.